\documentclass{article}

\usepackage[final]{ewrl_2023}

\usepackage[utf8]{inputenc} %
\usepackage[T1]{fontenc}    %
\usepackage[backref=page]{hyperref}       %
\usepackage{url}            %
\usepackage{booktabs}       %
\usepackage{amsfonts}       %
\usepackage{nicefrac}       %
\usepackage{microtype}      %
\usepackage{xcolor}         %
\usepackage{float}

\usepackage{amsthm}
\usepackage{amsmath}
\theoremstyle{theorem}
\newtheorem{theorem}{Definition}
\newtheorem{corollary}{Corollary}[theorem]

\DeclareMathOperator*{\argmax}{arg\,max}

\usepackage{graphicx}
\usepackage{wrapfig}
\usepackage{caption}
\usepackage{subcaption}

\usepackage{lipsum}

\usepackage{tikz}
\usepackage{import}
\usetikzlibrary{quotes,arrows.meta}
\usetikzlibrary{positioning}

\def\edgecolor{rgb:blue,4;red,1;green,4;black,3}
\newcommand{\midarrow}{\tikz \draw[-Stealth,line width =0.8mm,draw=\edgecolor] (-0.3,0) -- ++(0.3,0);}

\usepackage{Ball}
\usepackage{Box}
\usepackage{RightBandedBox}
\usetikzlibrary{positioning}
\usetikzlibrary{3d} %

\def\ConvColor{rgb:yellow,5;red,2.5;white,5}
\def\ConvReluColor{rgb:yellow,5;red,5;white,5}

\def\FcColor{rgb:blue,5;red,2.5;white,5}
\def\FcReluColor{rgb:blue,5;red,5;white,4}

\title{The Role of Diverse Replay for Generalisation in Reinforcement Learning}

\author{%
  Max Weltevrede \\
  Delft University of Technology \\
  Delft, 2628 XE, The Netherlands \\
  \texttt{m.r.weltevrede@tudelft.nl}
   \And
   Matthijs T.J. Spaan \\
  Delft University of Technology \\
  Delft, 2628 XE, The Netherlands 
   \AND
   Wendelin B\"ohmer \\
  Delft University of Technology \\
  Delft, 2628 XE, The Netherlands 
}

\begin{document}

\maketitle

\begin{abstract}
In reinforcement learning (RL), key components of many algorithms are the exploration strategy and replay buffer. These strategies regulate what environment data is collected and trained on and have been extensively studied in the RL literature. In this paper, we investigate the impact of these components in the context of generalisation in multi-task RL. We investigate the hypothesis that collecting and training on more diverse data from the training environments will improve zero-shot generalisation to new tasks. We motivate mathematically and show empirically that generalisation to tasks that are ``reachable'' during training is improved by increasing the diversity of transitions in the replay buffer. Furthermore, we show empirically that this same strategy also shows improvement for generalisation to similar but ``unreachable'' tasks which could be due to improved generalisation of the learned latent representations. 
\end{abstract}

\section{Introduction}
\begin{wrapfigure}{R}{0.5\textwidth}
\centering
\includegraphics[width=.49\textwidth]{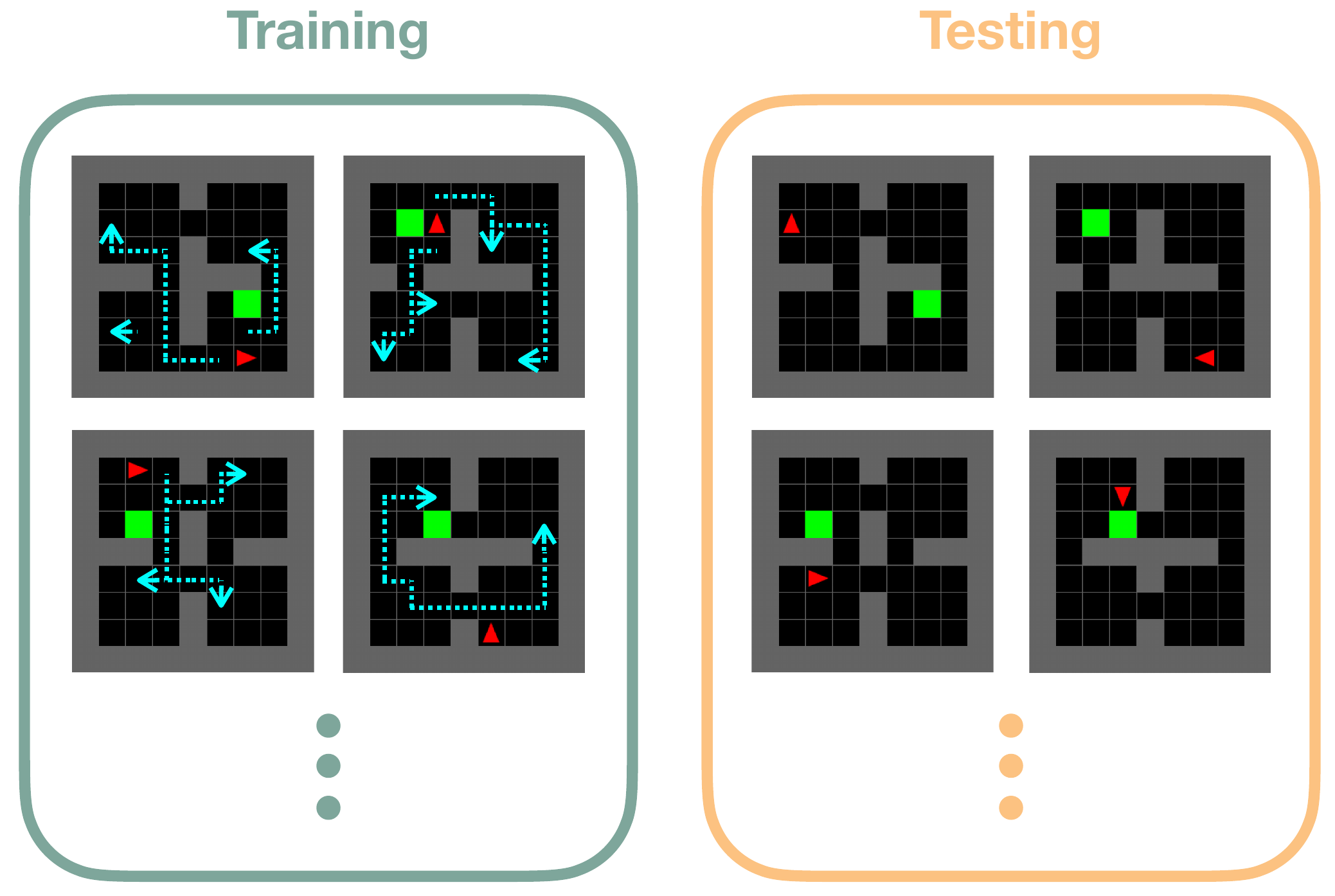}
\caption{Example of a generalisation problem to various 4-room tasks. With sufficient exploration (cyan) in the training tasks, the agent can encounter states that will likely help it generalise to the testing tasks.}
\label{fig:motivating}
\end{wrapfigure}

An important aspect of reinforcement learning research is investigating methods that can generalise from the training conditions to various new and unseen deployment scenarios. In this paper, we refer to generalisation as the goal of transferring a policy trained on a set of training tasks to a set of testing tasks. Popular approaches in the current literature try to improve generalisation performance by increasing the similarity or explicitly tackling the differences between the training and testing tasks \citep{kirk_survey_2023}. What these methods have in common is that they focus on the differences between the distributions of tasks, rather than the specific data (states and actions) encountered during those tasks. 

Key components of many reinforcement learning algorithms are the strategies to collect, store and sample data on which to train. For this reason, extensive literature exists on the problems of exploration (collect) \citep{amin_survey_2021}, replay buffers (store) \citep{isele_selective_2018, bruin_improved_2016} and replay sampling strategies (sample) \citep{schaul_prioritized_2016, zhang_deeper_2017, andrychowicz_hindsight_2017}. It is common in generalisation approaches to treat the exploration and replay storing and sampling strategies in the same way one would in singleton RL: only explore and replay new transitions to the extent necessary to learn the current task. 

In this paper, we investigate the role of the exploration strategy and the replay buffer on generalisation performance beyond the singleton perspective. We theorise that a more diverse strategy of collecting and sampling training data will improve performance in many generalisation scenarios (see Figure \ref{fig:motivating}). Our key contributions are the following:
\begin{itemize}
    \item We define a measure of \emph{reachability} that will allow us to prove that more diverse exploration and replay strategies lead to improved generalisation performance under certain restrictive conditions.
    \item We empirically show that generalisation to \emph{reachable} tasks benefits from more diverse exploration and replay strategies.
    \item We show empirically that generalisation to similar but \emph{unreachable} tasks also benefits from more diverse exploration and replay strategies.
    \item We analyse the latent representations induced by more diverse exploration and replay strategies and find they are likely responsible for the improved generalisation performance.
\end{itemize}

\section{Background}
The goal of reinforcement learning is to optimise decision-making in a Markov decision process (MDP). A Markov decision process is often denoted as a 6-tuple $\mathcal{M} = (S, A, T, R, p_0, \gamma)$ where $S$ is a set of states called the state space, $A$ a set of actions called the action space, $T: S \times A \to \mathcal{P}(S)$ a transition probability function (where $ \mathcal{P}(\mathcal{X})$ denotes the set of probability distributions over set $\mathcal{X}$), $R: S \times A \to \mathbb{R}$ a reward function, $p_0: \mathcal{P}(S)$ a distribution of initial states and $\gamma \in [0, 1)$ a discount factor. In reinforcement learning, the goal is to find the policy function $\pi: S \to \mathcal{P}(A)$ that maximises in expectation the discounted sum of future rewards: $\pi^* := \argmax_{\pi} \mathbb{E}_{\pi}\big[\sum_{t=0}^{\infty} \gamma^t r_t \big]$ where $\pi^*$ is the optimal policy and $\mathbb{E}_{\pi}$ denotes the expectation over the Markov chain $(s_0, a_0, r_0, s_1, a_1, r_1, ...)$ induced by the policy $\pi$, initial state distribution $p_0$, transition function $T$ and reward function $R$ \citep{akshay_steady-state_2013}. The steady-state distribution $\rho^\pi : \mathcal{P}(S)$ of the Markov chain induced by policy $\pi$ in MDP $\mathcal{M}$ defines the proportion of time spent in each state as the number of transitions within $\mathcal{M}$ goes to infinity. This distribution can be proven to be unique under certain conditions on the policy $\pi$ and MDP $\mathcal{M}$ \citep{konstantopoulos_markov_2009}. 

In Q-learning the goal is to learn the optimal Q-function $Q^*(s,a) := \mathbb{E}_{\pi^*}\big[\sum_{t=0}^{\infty} \gamma^t r_t \big|_{a_0 = a}^{s_0 = s} \big]$ which represents the expected future discounted return given a current state $s$ and action $a$. Deep Q learning \citep[DQN,][]{mnih_human-level_2015} is a popular deep reinforcement learning algorithm that trains a neural network $q_\phi$ to satisfy the optimal Bellman equation $Q^*(s,a) = R(s,a) + \gamma  \max_{a'} Q^*(s',a')$. It achieves this by minimising the mean squared error between the left-hand side and right-hand side of the optimal Bellman equation $\min_\phi \mathbb{E}\big[(r + \gamma \max_{a'}q_{\phi'}(s',a') - q_\phi(s,a))^2 \big| \langle s,a,r,s' \rangle \sim \mathcal{D}\big]$, where $q_{\phi'}$ is a target network which is periodically updated with the parameters of $q_{\phi}$ and $\mathcal{D}$ is a dataset of transitions called the replay buffer. The function $q_\phi$ is trained \emph{online} by interleaving gradient steps with data collection for the replay buffer. Typically, the replay buffer has fixed size and stores transitions collected during training in a first-in-first-out (FIFO) principle. Transitions are commonly collected with an $\epsilon$-greedy exploration policy: $\pi_{\epsilon}(a|s) = (1-\epsilon) \pi(a|s) + \epsilon \frac{1}{|A|}$, which selects random actions with probability $\epsilon$ or a `greedy' action with probability $1-\epsilon$ (according to $\pi(a|s) = 1$ if $a = \argmax_{a'} q_{\phi}(s,a')$ and $\pi(a|s) = 0$ otherwise). The inherent generalising capabilities of the neural network in DQN allow the agent to learn even in continuous state spaces (where it is impossible to collect experience for every possible state).

\subsection{Contextual Markov Decision Process}
A contextual MDP \citep[CMDP,][]{hallak_contextual_2015} is a specific type of MDP where the state space $S$ is factored into an underlying state space $S'$ and a \emph{context space} $C$: $S = S' \times C$. The context $c \in C$  is drawn at the start of an episode and fixed thereafter. The context can be thought of as defining the task an agent has to solve in a multi-task setup and we will refer to contexts as tasks from now on. In this paper, we will assume the task $c$ is fully observable, which means the state $s_0 \in S$ at the start of an episode fully determines the task the agent has to solve that episode. An example can be found in Figure \ref{fig:motivating}  where the agent's starting location and orientation, the goal location and the location of the doorways (referred to as the 4-room \emph{topology}) fully determines the task to be solved in that episode. 

\subsection{Zero-shot Policy Transfer}
\label{sec:zspt}
It is possible to define a generalisation setting by defining an MDP $\mathcal{M}|_{C}$ with a task space $C$ and task sets $C_{train}, C_{test} \subseteq C$ sampled during training and testing. For example, in a typical generalisation setting the agent is tested in the MDP $\mathcal{M}|_{C_{test}}$ on tasks $C_{test}$ that are unseen (held-out) during training ($C_{train} \cup C_{test} = \emptyset$). 

In particular, we consider in this paper the zero-shot policy transfer (ZSPT) problem class as defined in \citet{kirk_survey_2023}. A ZSPT problem is defined through a choice of training and testing task sets $C_{train}, C_{test}$ with the objective of maximising return in the testing MDP $\mathcal{M}|_{C_{test}}$, whilst only being allowed to train on $\mathcal{M}|_{C_{train}}$. The agent is expected to perform \emph{zero-shot} generalisation in the testing tasks, so without a fine-tuning or adaptation period. We assume the task $c$ is fully observable at the start of an episode, and therefore determined by the starting state $s_0 \in S$, we will refer to the task and starting state interchangeably ($\mathcal{M}|_{C_{train}}$ becomes $\mathcal{M}|_{S_0^{train}}$ where $S_0^{train}$ is the set of starting states $s_0 \in S_0^{train}$ that correspond with $C_{train}$).

\section{Related Work}
Zero-shot generalisation in contextual MDPs is a well-studied area of research with a diverse range of different approaches. One branch of research is focused on increasing the similarity between the training and testing task sets. Some examples of this are the use of (visual) data augmentations \citep{raileanu_automatic_2021, lee_network_2020, zhou_domain_2021} or (dynamical) domain randomisation \citep{tobin_domain_2017, sadeghi_cad2rl_2017, peng_sim--real_2018} to increase the complexity of the training tasks, with the aim to subsume the testing tasks in this increased training complexity. Another branch of research is focused on explicitly handling the (expected) differences between training and testing. This includes approaches that use inductive biases \citep{raileanu_automatic_2021, higgins_darla_2017, zambaldi_relational_2018, zambaldi_deep_2019, kansky_schema_2017, wang_unsupervised_2021, tang_neuroevolution_2020, tang_sensory_2021} or non-RL specific regularisation \citep{cobbe_quantifying_2019, ada_generalization_2019, tishby_deep_2015, igl_generalization_2019, lu_dynamics_2020, eysenbach_robust_2021} to overcome the challenges induced by the variations between training and testing. 

Multi-task reinforcement learning can sometimes refer to several agents jointly learning to solve a set of tasks whilst sharing some form of knowledge amongst the agents \citep{vithayathil_varghese_survey_2020}. However, in this paper, we exclusively refer to multi-task reinforcement learning as a single agent learning to solve multiple tasks in a contextual MDP.

All the above-mentioned works use techniques to improve generalisation that are not necessarily specific to RL. For example, most of the work applies ideas from supervised learning focused on the differences between the training and testing datasets to the differences between the training and testing \emph{task sets}, rather than the data collected from those tasks. Others \citep{igl_transient_2021, lyle_learning_2022} do look into RL-specific problems with generalisation but are mostly focused on the effects non-stationary data or temporal difference update rules have on the generalisation capacity of neural networks. The work most similar to ours is a workshop paper \citep{jiang_uncertainty-driven_2022} that proposes a new exploration algorithm motivated by a similar argument as used in our paper. Our work differs in that we don't propose a new exploration method, but instead investigate how and why exploration and replay buffer strategies affect generalisation performance in certain types of contextual MDPs. As such, in Section \ref{sec:latent} we observe primarily a representation learning effect which seems in contrast to some of the results in \citep{jiang_uncertainty-driven_2022} that suggest RL-specific effects beyond representation learning. Furthermore, we make a clear distinction between reachable and unreachable generalisation as two distinct scenarios requiring separate theoretical motivations.

\section{Method}
In the multi-task RL literature, exploration and experience replay is often treated as one would do in singleton RL: only explore and replay new transitions enough to learn the current task. In this section, we will define a generalisation setting to \emph{reachable states} and use it to prove that the singleton perspective on exploration and experience replay can be sub-optimal for generalisation in multi-task RL.

\begin{wrapfigure}{R}{.35\textwidth}
\centering
\includegraphics[width=.34\textwidth]{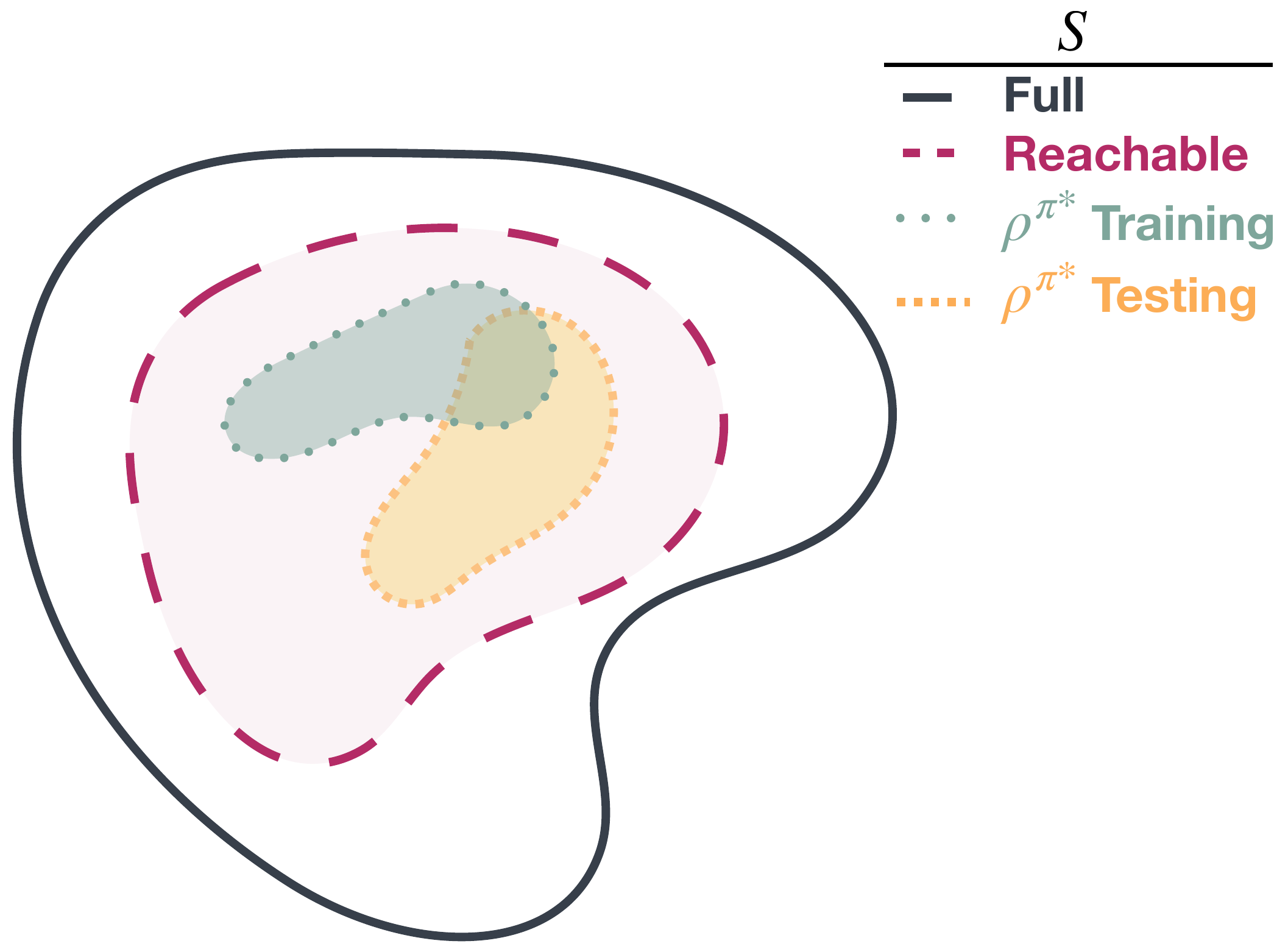}
\caption{Illustrative example of the state space covered by the set of reachable states $S_r(\mathcal{M}|_{S_0^{train}})$ and the steady state distributions $\rho^{\pi^*}$ of the optimal policy during training and testing in a ZSPT-R problem. It illustrates how a broader optimal coverage of the reachable state space during training can improve test time performance.}
\label{fig:coverage}
\vspace{-16mm}
\end{wrapfigure}

We define a reachable state $s_r$ as a state for which there exists a policy whose probability of encountering $s_r$ during training is non-zero. This gives us the following definition of a set of reachable states $S_r$ in an MDP $\mathcal{M}|_{S_0^{train}}$.\footnote{Similar to a definition of reachability for Markov chains in \citet{velasquez_optimal_2023}}
\begin{theorem}
\label{thrm:reachability}
    The set of reachable states $S_r(\mathcal{M}|_{S_0^{train}})$ consists of all states $s_r$ for which there exists a sequence of actions that give a non-zero probability of ending up in $s_r$ when performed in the MDP $\mathcal{M}|_{S_0^{train}}$.
\end{theorem}
\begin{corollary}
\label{cor:closed}
    Any state $s'$ that is reachable from a state $s \in S_r(\mathcal{M}|_{S_0^{train}})$ in the reachable set, has to be itself in the reachable set: $s' \in S_r(\mathcal{M}|_{S_0^{train}})$.
\end{corollary}
Corollary \ref{cor:closed} entails that you cannot leave the reachable set $S_r(\mathcal{M}|_{S_0^{train}})$ through interaction with the environment.

Using this definition of reachable states we can define a particular instance of the ZSPT problem.
We define \emph{ZSPT to reachable states} (ZSPT-R) as a ZSPT problem where the initial states during testing $S_0^{test}$ are part of the set of reachable states during training $S_0^{test} \subset S_r(\mathcal{M}|_{S_0^{train}})$. This particular instance of the ZSPT problem has an interesting property.
\begin{corollary}
    An optimal policy $\pi$ that achieves maximal return from any state in the reachable state space $S_r(\mathcal{M}|_{S_0^{train}})$, will have optimal performance in the ZSPT-R problem setting. 
\end{corollary}
Recall that performance in the ZSPT problem is defined as the performance in the testing MDP $\mathcal{M}|_{S_0^{test}}$, which in the case of ZSPT-R, has a state space that consists only of reachable states (due to Corollary \ref{cor:closed}) It follows naturally that an optimal policy on the entire reachable state space $S_r(\mathcal{M}|_{S_0^{train}})$ also has to  be optimal in $\mathcal{M}|_{S_0^{test}}$. 

So, in ZSPT-R it is better to explore and replay transitions from all over the reachable state space: even though some states might be irrelevant during training (not part of the steady-state distribution of the optimal policy), they could be crucial during testing (see Figure \ref{fig:coverage}). This is in contrast with singleton RL, where it can be (correctly) assumed that states not encountered by the optimal policy during training, will not be encountered by the optimal policy during testing.

The above holds for an idealised scenario where it is possible to learn an exact optimal policy for every reachable state. However, with function approximation in large state-action spaces, the exact optimal solution might not always be achievable. Nevertheless, a more diverse exploration and replay of the reachable state space will likely improve generalisation performance. 

This insight does not directly carry over to the general ZSPT problem: if testing is no longer constrained to reachable states $S_r(\mathcal{M}|_{S_0^{train}})$, then training on a larger part of the reachable state space might not improve test performance. However, with the inherent generalisation capabilities of neural networks, we still expect a more diverse replay buffer to benefit generalisation to unreachable (but similar) states through learning better generalising latent representations.

\section{Experiments}
In this section, we empirically evaluate the role of a diverse replay buffer on generalisation to new starting states. We first validate that increased diversity of the replay buffer (through exploration and replay buffer parameters) in the function approximation setting will lead to a policy that is optimal on a larger part of the reachable state space. We then evaluate the effect a more diverse replay buffer has on generalisation performance to both reachable and unreachable starting states. Lastly, we analyse the latent representations of the agents trained with different levels of replay buffer diversity.

\subsection{4-Room Grid World}
Our investigation is performed in a small 4-room grid world (Figure \ref{fig:motivating}) adapted from the \texttt{FourRooms} environment from the MiniGrid benchmark \citep{chevalier-boisvert_minimalistic_2018}. The state space (including the task space) is fully observable and has three actions: move forward, turn left, and turn right.  Between training and testing the only differences are the starting states $s_0$ (or equivalently, tasks), which vary along the following dimensions: the 4-room topology, the starting location of the agent, the direction the agent is facing and the location of the goal. During training, the agent will encounter a fixed number of different start states and will be tested on a number of unseen starting states (generated from the same distribution). See Appendix \ref{app:env} for more details on the environment.

In this environment, reachability is regulated through variations in the goal location and topology. If two states share their topology and goal location, then they are both reachable from one another. Conversely, if two states differ in either the topology or goal location, they are unreachable. 

\subsection{Policy Optimality in Reachable Space}
In this experiment, training is done on a fixed set of 40 starting states which differ in their agent location, agent direction, goal location and topology. The agent is trained using DQN with linearly decaying $\epsilon$-greedy exploration and a FIFO replay buffer. In order to increase the diversity of transitions in our replay buffer, we vary the amount of exploration performed by varying the duration over which $\epsilon$ is linearly decayed. Note that increasing the amount of exploration does not necessarily lead to a consistently more diverse replay buffer, as the more diverse exploratory data might drop out of the first-in-first-out buffer. To avoid this, we use a replay buffer size equal to the total length of training (500.000 environment steps). Therefore, the replay buffer never drops any experiences and varying the amount of exploration induces different ratios of diversity in the buffer. See Appendix \ref{app:exp} for more training details.

\begin{figure}[h]
\begin{center}
    \begin{subfigure}[b]{0.32\textwidth}
        \centering
        \includegraphics[width=1\textwidth]{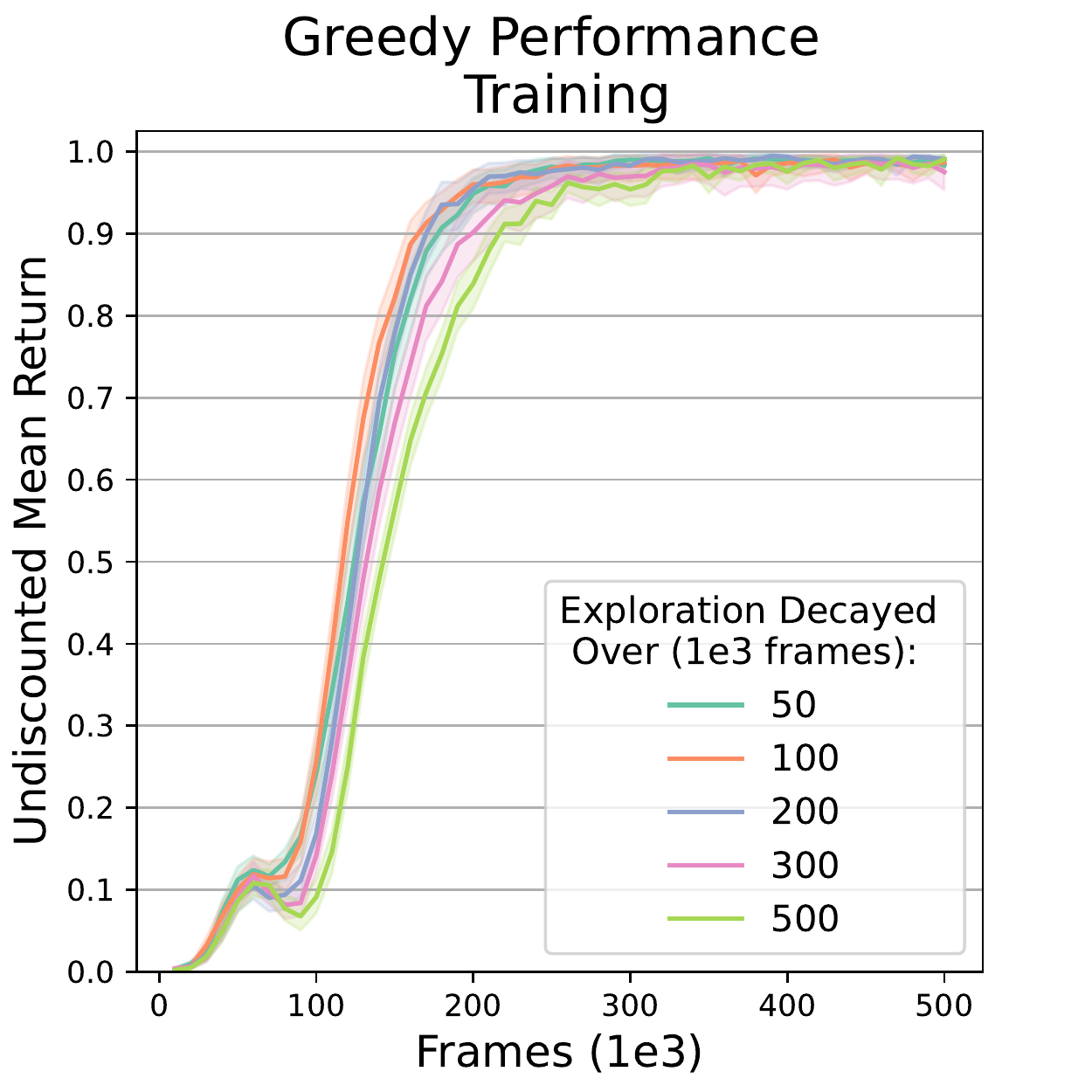}
        \caption{Training performance}
        \label{fig:exp1_train}
    \end{subfigure}
    \hfill
    \begin{subfigure}[b]{0.32\textwidth}
        \centering
        \includegraphics[width=1\textwidth]{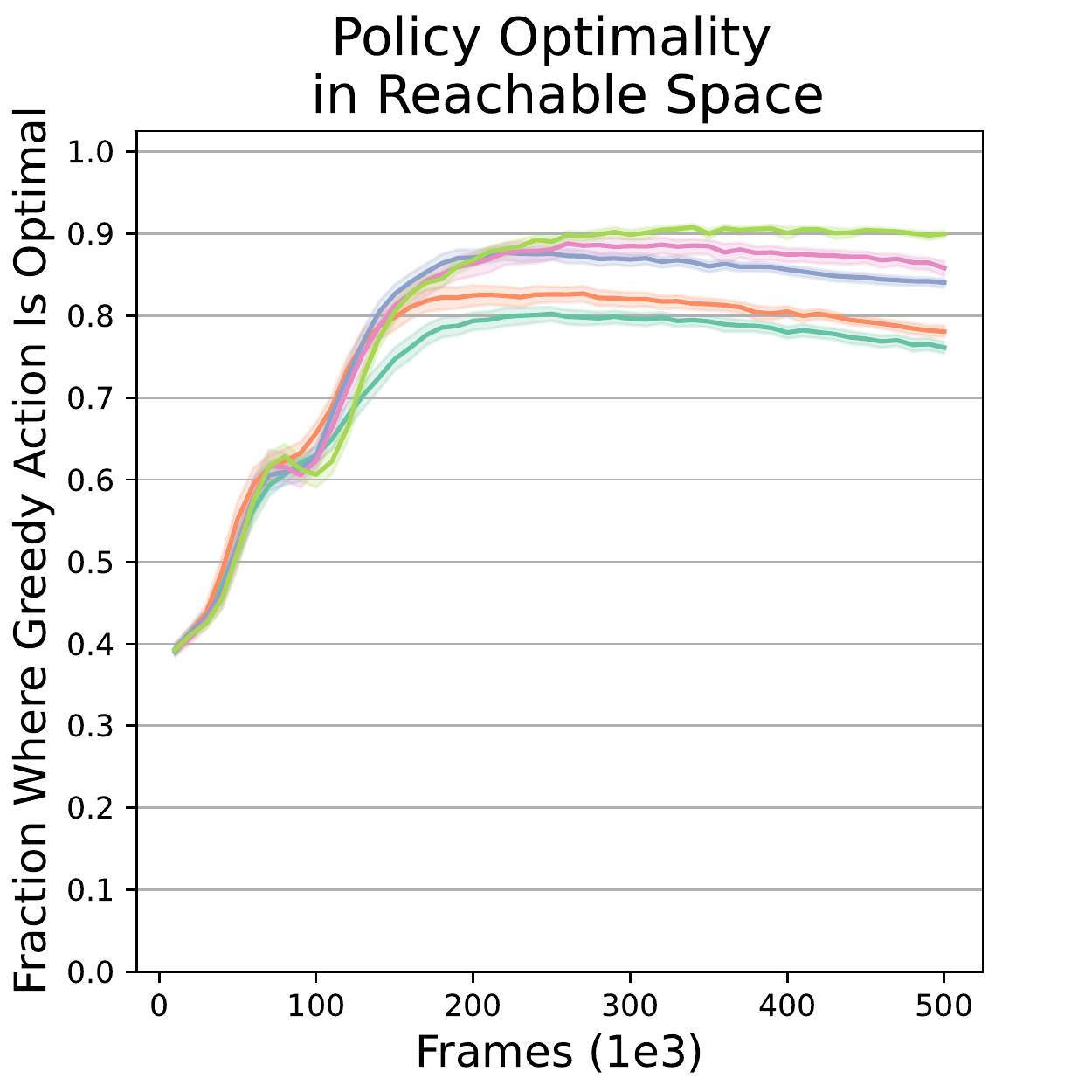}
        \caption{Policy Optimality}
        \label{fig:exp1_coverage}
    \end{subfigure}
    \hfill
    \begin{subfigure}[b]{0.32\textwidth}
        \centering
        \includegraphics[width=1\textwidth]{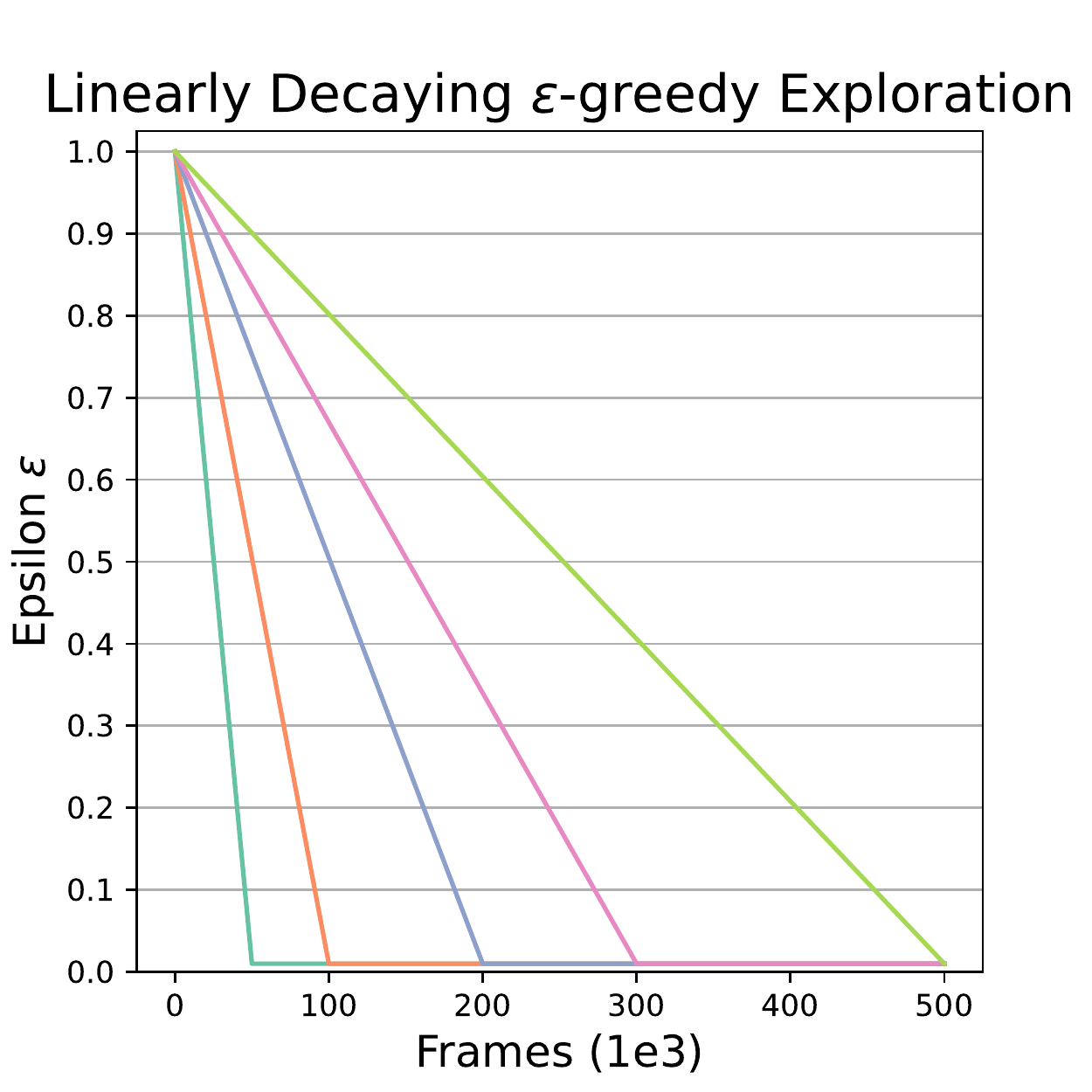}
        \caption{Linearly decaying $\epsilon$-greedy }
    \end{subfigure}
\end{center}
\caption{Left: the mean undiscounted return during training when following the greedy policy $\pi(s) = \argmax_a' Q(s,a')$. Middle: fraction of reachable space where the greedy policy is optimal. Right: $\epsilon$ in $\epsilon$-greedy exploration as it is decayed during training. Different degrees of replay buffer diversity are compared by using a sufficiently large replay buffer and varying the duration over which exploration is decayed. Shown are the mean and a 95\% confidence interval over 100 seeds. }
\label{fig:exp1}
\end{figure}

We compare five different linearly decaying $\epsilon-$greedy strategies (decayed over the first 50.000, 100.000, 200.000, 300.000 and 500.000 environment steps) and how they affect performance of the greedy policy $\pi(s) = \argmax_a' Q(s,a')$ during training and the fraction of reachable space for which our greedy policy is optimal. The results can be found in Figure \ref{fig:exp1}. Comparing Figures \ref{fig:exp1_train} with \ref{fig:exp1_coverage} shows a clear trend that more exploration, and therefore more diverse data in the replay buffer, leads to a higher fraction of reachable state space where the policy is optimal whilst training performance is mostly unaffected.

\subsection{Generalisation To Unseen Starting States}
\label{sec:gen}
For evaluating the generalisation capabilities, the agent is trained on the same 40 starting states as the previous experiment and tested on 40 new and unseen starting states. Two testing sets are constructed to evaluate the agent's performance on both reachable and unreachable starting states. 

The 100\% reachable test set is constructed by taking every training start state and changing only the agent location and agent direction (keeping the topology and goal location the same). The 0\% reachable (unreachable) test set is constructed by taking every start state in the 100\% reachable test set and changing the topology to one not encountered during training (keeping the agent location, direction and goal location the same as the 100\% reachable set). 

\begin{figure}[h]
\begin{center}
    \begin{subfigure}[b]{0.32\textwidth}
        \centering
        \includegraphics[width=1\textwidth]{images/train_transparent.pdf}
        \caption{Training performance}
        \label{fig:exp2_train}
    \end{subfigure}
    \hfill
    \begin{subfigure}[b]{0.32\textwidth}
        \centering
        \includegraphics[width=1\textwidth]{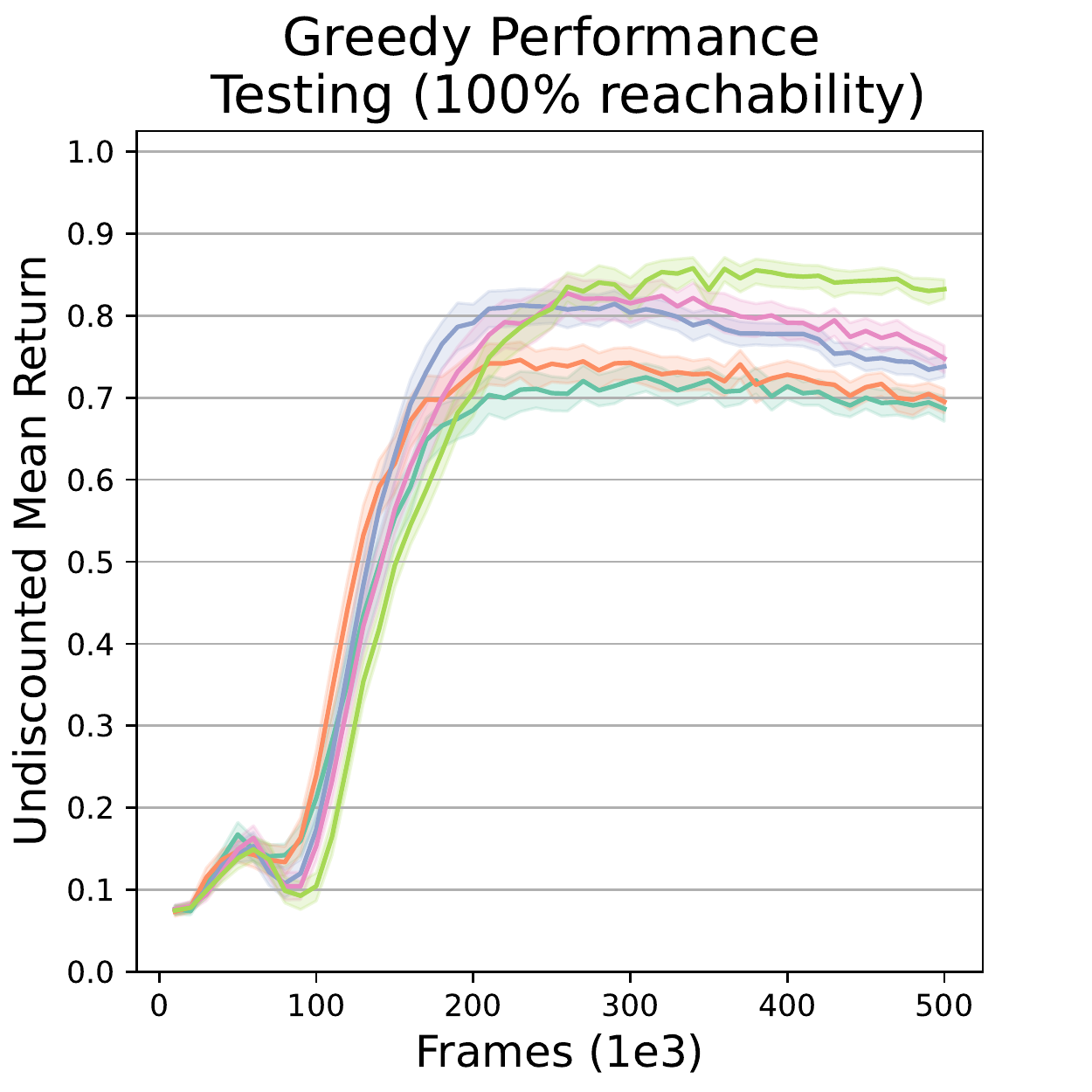}
        \caption{Testing performance}
        \label{fig:exp2_test100}
    \end{subfigure}
    \hfill
    \begin{subfigure}[b]{0.32\textwidth}
        \centering
        \includegraphics[width=1\textwidth]{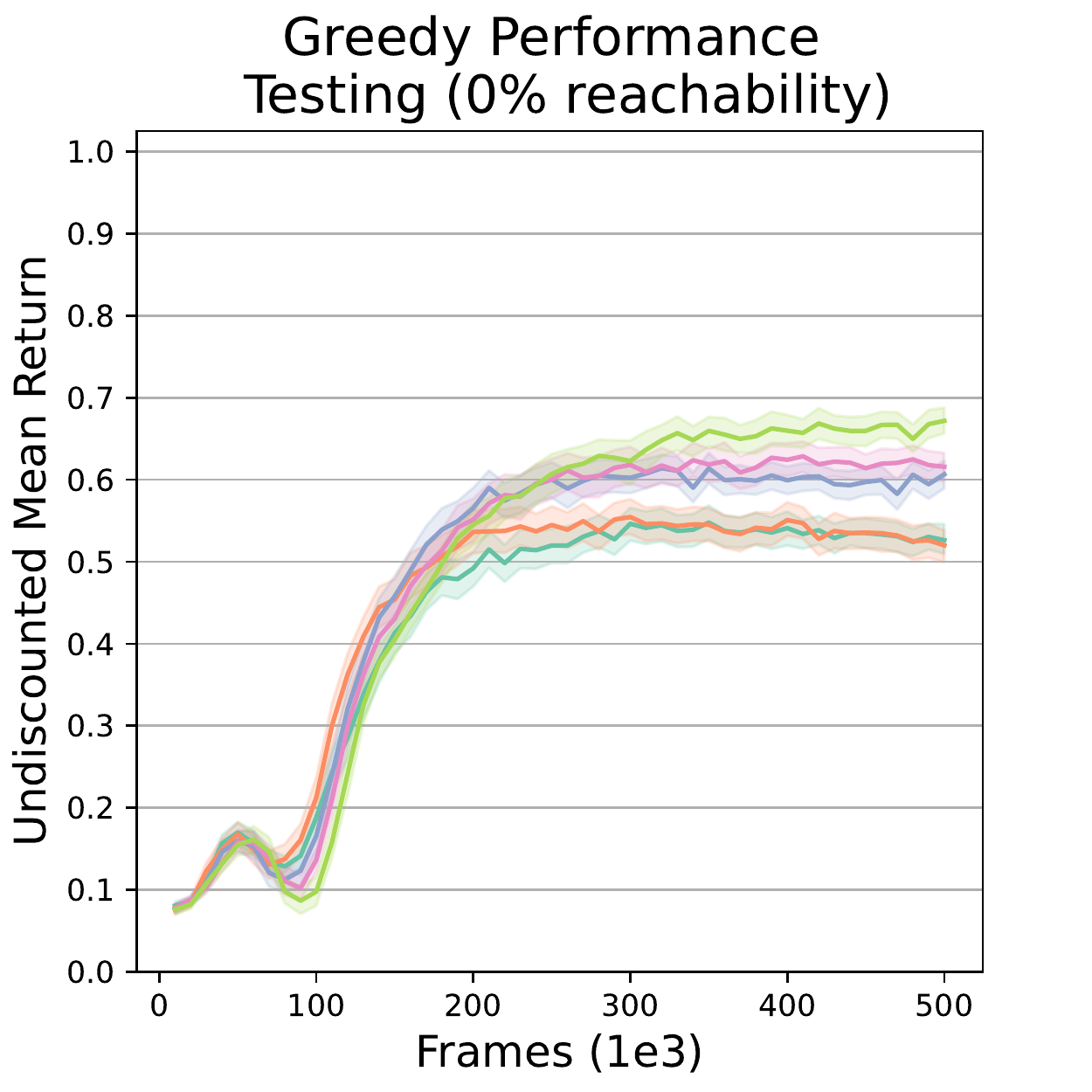}
        \caption{Testing performance}
        \label{fig:exp2_test0}
    \end{subfigure}
\end{center}
\caption{The mean undiscounted return when following the greedy policy $\pi(s) = \argmax_a' Q(s,a')$.  Different degrees of replay buffer diversity are compared by using a sufficiently large replay buffer and varying the duration over which exploration is decayed. Left: during training. Middle: when testing on a 100\% reachable test set. Right: when testing on a 0\% reachable test set. Shown are the mean and a 95\% confidence interval over 100 seeds. }
\label{fig:exp2}
\end{figure}

Figure \ref{fig:exp2} shows the greedy performance during training, testing on 100\% reachable start states and testing on 0\% reachable start states. Figure \ref{fig:exp2_train} shows the same training performance as can be found in Figure \ref{fig:exp1_train}. Comparing Figure \ref{fig:exp2_train} with \ref{fig:exp2_test100}, there is a clear trend that a more diverse replay buffer leads to better generalisation to reachable starting states (higher test performance whilst training performance is largely unaffected). Comparing this with Figure \ref{fig:exp1_coverage} shows a correlation between a higher fraction of reachable space where the policy is optimal and higher generalisation performance to reachable states. Figure \ref{fig:exp2_test0} shows that this trend even holds for generalisation to unreachable starting states.

\subsection{Uniform Replay over State-action Space}
\label{sec:uniform}
In order to further solidify the role of a diverse replay buffer, the experiment is repeated but with a \emph{$S,A$-uniform} replay buffer. An $S,A$-uniform replay buffer samples its mini-batches uniformly over the state-action space $(S \times A)_\mathcal{D} \subseteq S \times A$ supported in the buffer $(S \times A)_\mathcal{D} = \{(s,a) \: | \: (s,a) \in \mathcal{D} \}$, rather than uniformly over the experiences $(s,a,r,s') \in \mathcal{D}$ in the buffer (which we call buffer-uniform here). The resulting agent is trained  (approximately) uniform over the reachable state-action space (assuming sufficient exploration that ensures every $(s,a)$ is in the buffer at least once). We observed the $S,A$-uniform buffer to train slightly slower and therefore compensated by increasing the total amount of training steps (1 million steps). 

\begin{figure}[h]
\begin{center}
    \includegraphics[width=1\textwidth]{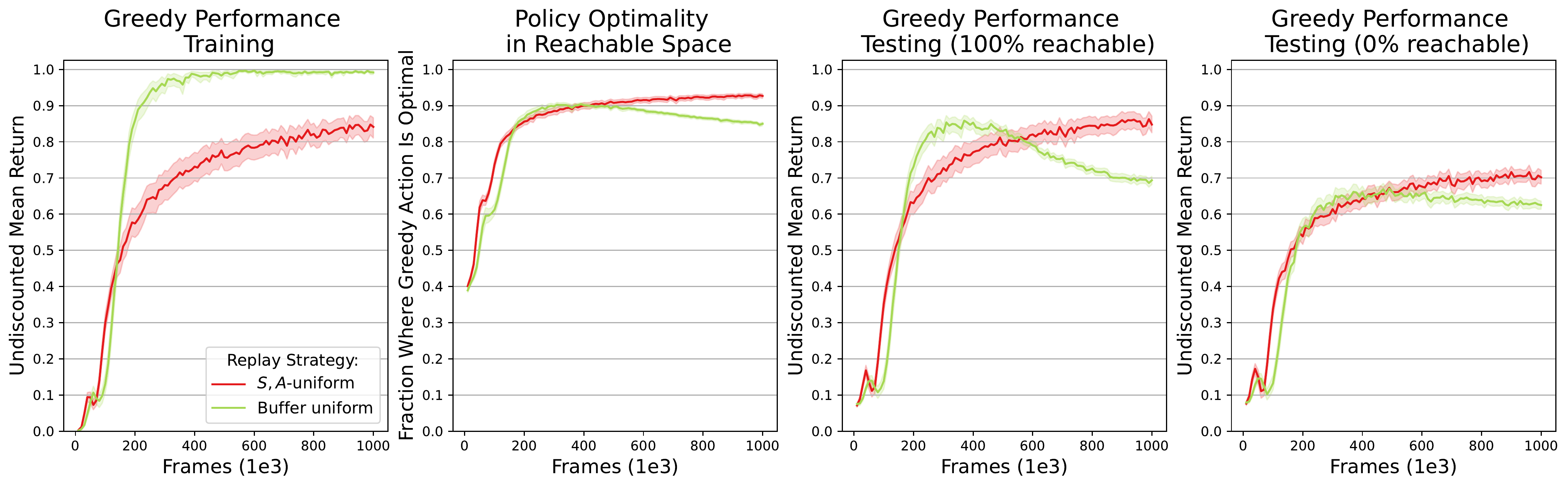}
    \caption{Performance of a uniform over state-action space ($S,A$-uniform) replay sampling strategy compared to the regular (buffer-uniform) replay sampling strategy. Both methods have $\epsilon$-greedy exploration decayed over 500.000 steps. Shown are the mean and 95\% confidence interval over 100 seeds.}
    \label{fig:max_cov}
\end{center}
\end{figure}

Figure \ref{fig:max_cov} shows the greedy performance of the $S,A$-uniform buffer during training and testing and its policy optimality in reachable space. For comparison, the best performing buffer-uniform replay buffer ($\epsilon$-greedy decayed over the first 500.000 steps) is shown for 1 million training steps. The exploration for the $S,A$-uniform buffer is also decayed over 500.000 steps to ensure sufficient coverage of reachable space and a fair comparison between the two methods. The figure shows the testing performance of the buffer-uniform replay buffer starts declining again after 500.000 steps. This is in line with our previous results as the fraction of diverse data sampled from the replay buffer starts declining after exploration has decayed.  In contrast, the $S,A$-uniform replay buffer does not suffer from this decline in diversity and instead keeps improving as training continues, outperforming the buffer-uniform buffer in everything other than training performance.

\subsection{Analysing Latent Representation}
\label{sec:latent}
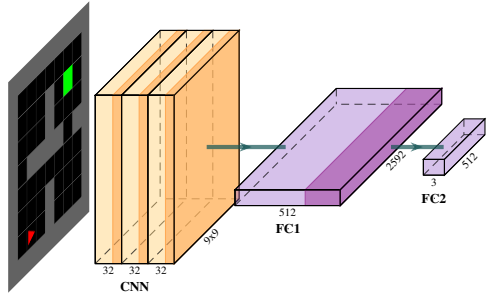
\begin{wrapfigure}{R}{0.5\textwidth}
\begin{tikzpicture}[scale=0.35, transform shape]
\tikzstyle{connection}=[ultra thick,every node/.style={sloped,allow upside down},draw=\edgecolor,opacity=0.7]
\tikzstyle{copyconnection}=[ultra thick,every node/.style={sloped,allow upside down},draw={rgb:blue,4;red,1;green,1;black,3},opacity=0.7]

\node[canvas is zy plane at x=0] (temp) at (-3,0,0) {\includegraphics[width=8cm,height=8cm]{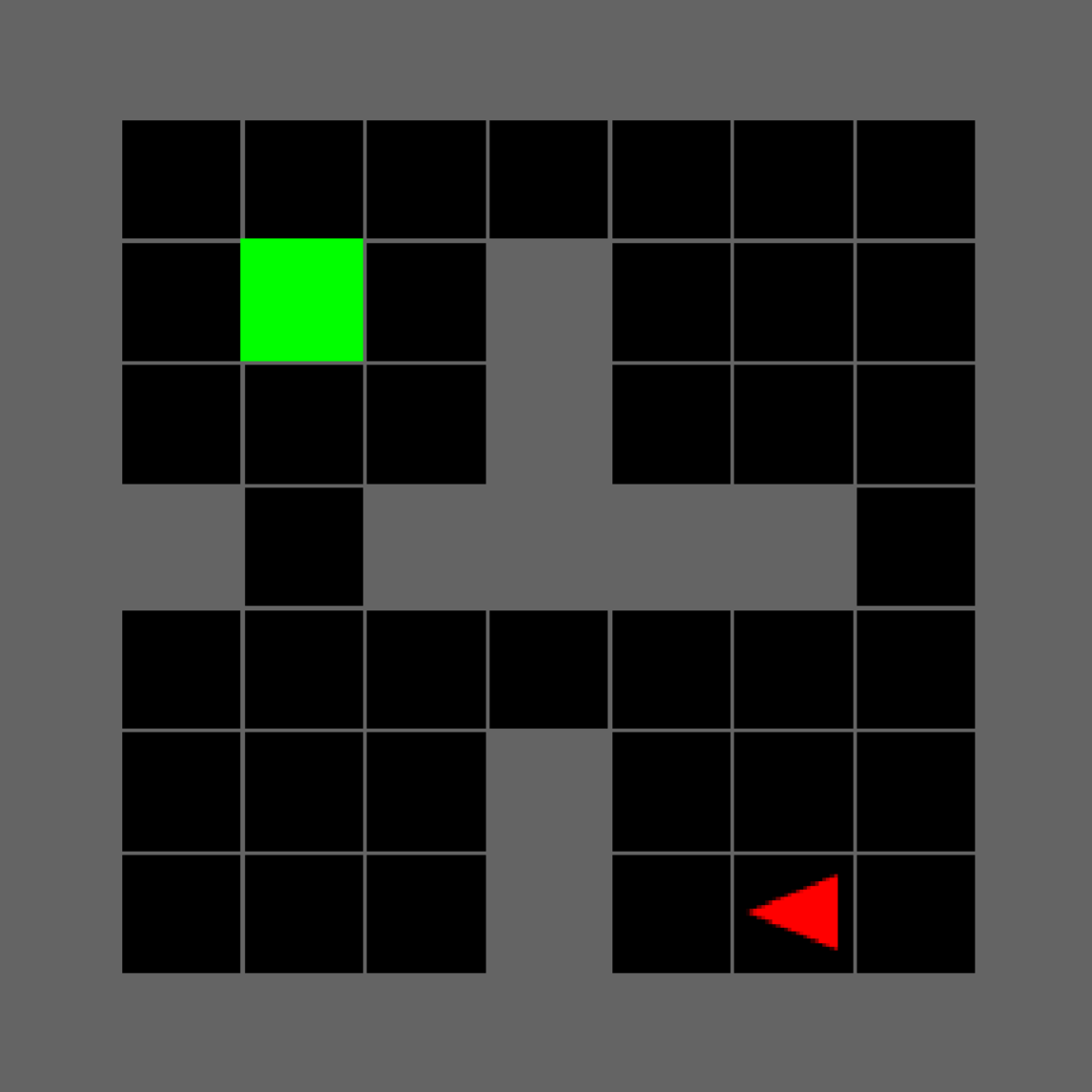}};

\pic[shift={ (0,0,0) }] at (0,0,0) 
    {RightBandedBox={
        name=conv1,
        caption=CNN,
        xlabel={{ 32, 32, 32 }},
        zlabel=9x9,
        fill=\ConvColor,
        bandfill=\ConvReluColor,
        height=32,
        width={ 5 , 5, 5 },
        depth=32
        }
    };
\pic[shift={(3,0,0)}] at (3,0,0) 
    {RightBandedBox={
        name=fc1,
        caption=FC1,
        xlabel={{512, }},
        zlabel=2592,
        fill=\FcColor,
        bandfill=\FcReluColor,
        height=3,
        width=20,
        depth=50
        }
    };
\pic[shift={(6,0,0)}] at (6,0,0) 
    {Box={
        name=fc2,
        caption=FC2,
        xlabel={{3, }},
        zlabel=512,
        fill=\FcColor,
        height=3,
        width=4,
        depth=20
        }
    };
\draw [connection]  (conv1-east)    -- node {\midarrow} (fc1-west);
\draw [connection]  (fc1-east)    -- node {\midarrow} (fc2-west);
\end{tikzpicture}
\caption{The architecture used in the experiments in Section \ref{sec:latent}. The dark-shaded regions indicate ReLU activation functions.}
\label{fig:arch}
\end{wrapfigure}

The following experiment examines what part of the network is responsible for the difference in (generalisation) performance of the agents trained in Section \ref{sec:gen}. We analyse the learned latent representations by fine-tuning them on the (narrow) steady-state distribution $\rho^{\pi^*}$ of the optimal policy during training. By evaluating on the entire reachable space and the test sets after fine-tuning we can evaluate whether the latent representations have learned to generalise or not. 

The experiment is structured as follows. In order to isolate the influence of the latent representation, we simplify the network by replacing the fully connected (FC) layers after layer FC1 with a linear probe \citep{alain_understanding_2016} (see Figure \ref{fig:arch}).  Parts of the network in Figure \ref{fig:arch} are frozen after the training done in Section \ref{sec:gen}. The downstream layers are reset and fine-tuned on narrow on-policy data collected by the converged (approximately optimal) agents. In order to avoid the challenges associated with offline RL \citep{levine_offline_2020}, we allow for a small amount of additional online data collection (up to 4.000 environment samples), which does not contain any explicit exploration. See Appendix \ref{app:latent} for more details. 

\begin{figure}[h]
\begin{center}
    \includegraphics[width=1\textwidth]{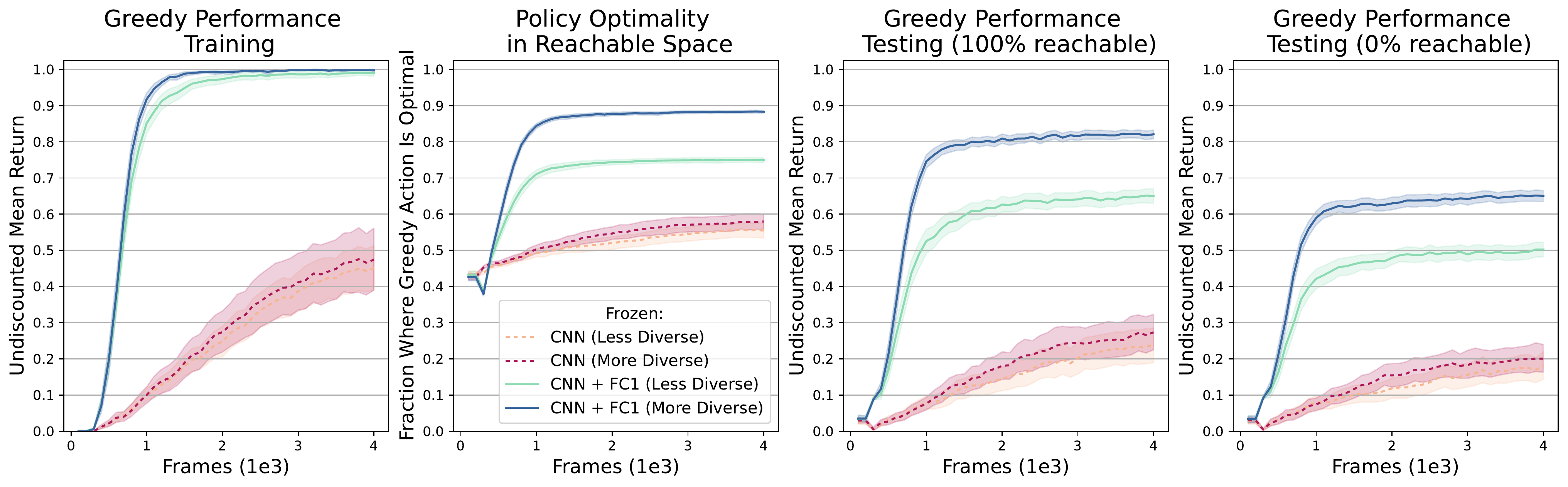}
    \caption{Fine-tuning performance with a frozen CNN or a frozen CNN and FC1. The less and more diverse frozen parameters are taken from the agents trained with $\epsilon$-greedy decayed after 50.000 and 500.000 steps respectively. Shown are the mean and 95\% confidence interval over 100 seeds.}
    \label{fig:fc_probe}
\end{center}
\end{figure}

Figure \ref{fig:fc_probe} shows the performance when freezing the CNN and FC1 layers and fine-tuning FC2. This is compared to fine-tuning FC1 and FC2 with only the CNN frozen. The frozen parameters are taken from agents trained with less diverse and more diverse buffers ($\epsilon$-greedy decayed over 50.000 and 500.000 steps respectively). The results show that the frozen CNN plus FC1, in contrast to only the frozen CNN, can quickly recover the training performance of the original agent. Moreover, it also recovers the generalisation performance, which depends on the diversity of the replay buffer of the original agent. 

This suggests that the FC1 layer after the CNN has learned a latent representation that generalises to a large part of the reachable and unreachable space, even if the following layer (FC2) is fine-tuned on very narrow transition data. Increasing the diversity of the training data on which the original agents were trained, seems to induce a better generalising latent representation, both in terms of reachable and unreachable generalisation. Furthermore, since fine-tuning recovers the original performance, it seems the difference in generalisation performance of the original agents in Section \ref{sec:gen} is due to the different latent representations learned by those agents. This seems in contrast to some of the results in \cite{jiang_uncertainty-driven_2022}, which suggest a component of the generalisation performance induced by more diverse training data is due to RL-specific effects beyond representation learning.

\section{Conclusion \& Future Work}
In this paper, we reasoned that being optimal in a larger part of the reachable state space will improve zero-shot policy transfer to reachable states.  We empirically showed this to hold in a 4-room grid world environment. We found a more diverse replay buffer to produce a policy that is optimal in a larger part of reachable space and has higher generalisation performance to reachable states.  Additionally, a more diverse replay buffer was empirically shown to lead to improved generalisation performance to similar (in-distribution) but unreachable states. Finally, we found that the latent representation induced by the fully connected layer following the CNN portion of the network generalises to reachable and unreachable states when fine-tuned on very narrow data. 

We believe this work demonstrates the non-trivial role of exploration and the replay buffer for generalisation in reinforcement learning. For now, this has been demonstrated in a relatively simple grid world environment. In future work, we would like to investigate whether the generalisation benefits will sustain when scaling up to more complex benchmarks.  Furthermore, we have an intuition on why a more diverse exploration and replay strategy would result in better generalisation performance to unreachable states. We believe it could be promising to formalise and test this intuition in follow-up experiments.

\newpage
\normalsize
\appendix
\begin{table}[h]
\centering
\begin{tabular}{@{}ll@{}}
\toprule
\textbf{Hyper-parameter}               & \textbf{Value}     \\ \midrule
Total timesteps                       & 500 000            \\
Vectorised environments               & 10                 \\ \midrule
\multicolumn{2}{c}{\textbf{DQN}}                           \\
Buffer size                           & 500 000            \\
Batch size                            & 256                \\
Discount factor $\gamma$              & 0.99               \\
Max. gradient norm                    & 1                  \\
Gradient steps                        & 1                  \\
Train frequency (steps)               & 10                 \\
Target update interval (steps)        & 10                 \\
Target soft update coefficient $\tau$ & 0.01               \\
Exploration initial $\epsilon$        & 1                  \\
Exploration final $\epsilon$          & 0.01               \\ \midrule
\multicolumn{2}{c}{\textbf{Adam}}                          \\
Learning rate                         & $1 \times 10^{-4}$ \\
Weight decay                          & $1 \times 10^{-5}$ \\ \midrule
\multicolumn{2}{c}{\textbf{CNN}}                           \\
Kernel size                           & 3                  \\
Stride                                & 1                  \\
Padding                               & 1                  \\
Padding mode                          & Circular           \\
Channels                              & 32                
\end{tabular}
\caption{Hyper-parameters used for the experiments}
\label{tab:parameters}
\end{table}

\section{4-Room Grid World}
\label{app:env}
The 4-room grid world used in our experiments is adapted from the \texttt{FourRooms} environment from the MiniGrid benchmark \citep{chevalier-boisvert_minimalistic_2018} and differs in certain ways from the default MiniGrid configuration. For one, the action space is reduced from the default seven actions (turn left, turn right, move forward, pick up an object, drop an object, toggle/activate an object, end episode) to just the first three actions (turn left, turn right, move forward). Also, the reward function is changed slightly to reward 1 for successfully reaching the goal and 0 otherwise (as opposed to the $1 - 0.9 * (\frac{\text{step count}}{\text{max steps}})$ given upon success by the default MiniGrid environment). Additionally, the size of the environment is reduced from the default 19 ($8 \times 8$ rooms) to 9 ($3 \times 3$ rooms).

Furthermore, the observation space is made fully observable and customised. Our agent receives a $4 \times 9 \times 9$ tensor that is centred around the agent's current location. The four binary-encoded channels contain the following information:
\begin{itemize}
    \item \textbf{Channel 0:} The location of the agent (always in the centre). 
    \item \textbf{Channel 1:} The hypothetical location where the agent would move to given the current direction it's facing (and ignoring any collisions with walls).
    \item \textbf{Channel 2:} The location of the walls. 
    \item \textbf{Channel 3:} The location of the goal. 
\end{itemize}

The implementation of the 4-room grid is also customised to allow for more control over the factors of variation (topology, agent location, agent direction, goal location) during the generation of a task. This acts functionally the same as the \texttt{ReseedWrapper} from MiniGrid except that it allows for more control and therefore easier design and construction of the training and testing sets.

\section{Experimental Details}
\label{app:exp}
We use a DQN implementation forked from the Stable-Baselines3 \citep{raffin_stable-baselines3_2021} repository and adapted to support double Q learning \citep{van_hasselt_deep_2015}. The network architecture consists of three convolutional layers followed by four fully connected layers with ReLU activation functions (except for the last layer). The hidden dimensions of the linear layers are $[512, 128, 64]$. A full list of parameters can be found in Table \ref{tab:parameters}. The experiments are performed on a single computer with a RTX 3070 GPU, 32 GB of memory and 12 cores (8 performance and 4 efficient). A single run of 500.000 steps had a runtime of roughly 8 minutes. 

\section{Analysing Latent Representation}
\label{app:latent}
In order to more accurately analyse the performance due to some latent part of the network, we follow a process akin to linear probing \citep{alain_understanding_2016}. For the agent with frozen CNN and FC1, the original remaining network (see above) is shortened by appending only a single linear layer (without activation function) to the frozen layers. We tried the same with the frozen CNN but found the single linear layer to not be sufficient to learn anything. Instead, we keep the fully connected layer FC1 followed by a single linear layer (but reset and fine-tune both). This way the network architectures are also identical between the two agents with differently frozen layers.  

To obtain the steady-state distribution $\rho^{\pi^*}$ of the optimal policy during training we simply perform greedy rollouts by the converged policies of our agents trained with $\epsilon$-greedy decayed over 500.000 steps. At the start of fine-tuning, the agent's buffer is initialised with the data collected from these rollouts. Fine-tuning is performed with the regular DQN algorithm but with certain hyper-parameters adjusted to speed up training and prohibit exploration. The adjusted hyper-parameters can be found in Table \ref{tab:adjusted}.

\begin{table}[h]
\centering
\begin{tabular}{@{}ll@{}}
\toprule
\textbf{Hyper-parameter}               & \textbf{Value}     \\ \midrule
Total timesteps                       & 4 000            \\
Vectorised environments               & 1                 \\ \midrule
\multicolumn{2}{c}{\textbf{DQN}}                           \\
Buffer size                           & 4 000            \\
Gradient steps                        & 10                  \\
Train frequency (steps)               & 1                 \\
Target update interval (steps)        & 1                 \\
Target soft update coefficient $\tau$ & 1               \\
Exploration initial $\epsilon$        & 0                  \\
Exploration final $\epsilon$          & 0               \\ \midrule        
\end{tabular}
\caption{Hyper-parameters used for the experiments in Section \ref{sec:latent}}
\label{tab:adjusted}
\end{table}

\subsection{Additional Experiments}
\label{app:small_network}
\begin{figure}[h]
\begin{center}
    \includegraphics[width=1\textwidth]{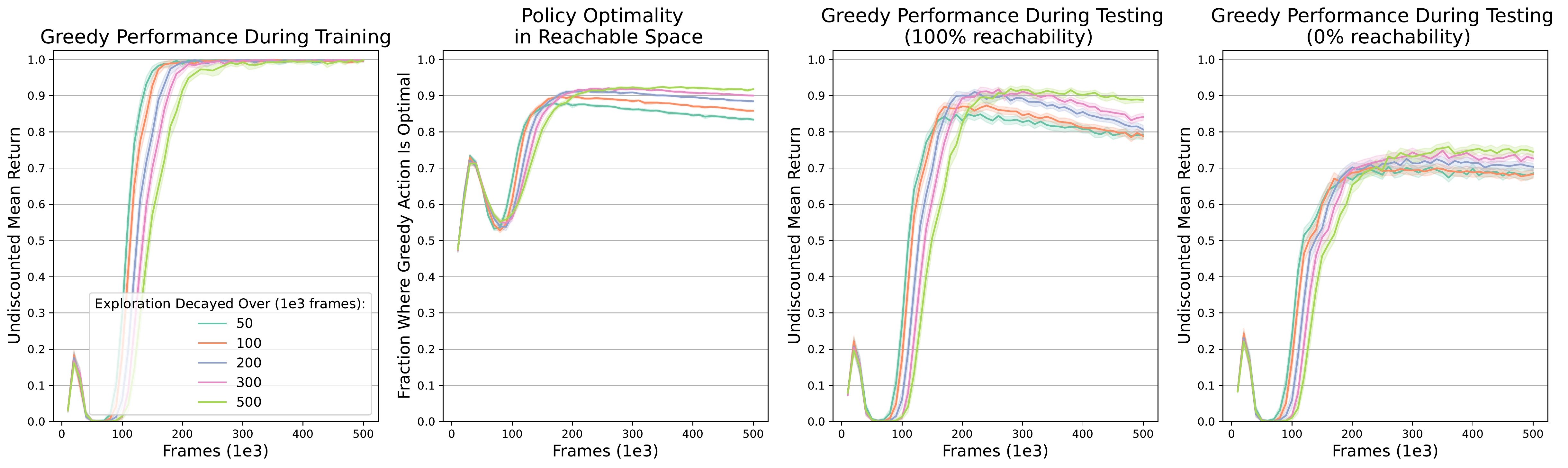}
    \caption{Performance when using a smaller network as shown in Figure \ref{fig:arch}. Shown are the mean and 95\% confidence interval over 100 seeds.}
    \label{fig:small_network}
\end{center}
\end{figure}

Since, the fine-tuning in Section \ref{sec:latent} recovered the original performance of the agents, it seems the network used in the experiments in sections \ref{sec:gen} and \ref{sec:uniform} was over-parametrised. For completeness, we reran the main experiment in Section \ref{sec:gen} but with the smaller network as shown in Figure \ref{fig:arch}. The results can be found in Figure \ref{fig:small_network}. 

The results show that the smaller network achieves a higher policy optimality in reachable space and a higher generalisation performance to reachable and unreachable states than the original network. Moreover, the difference between the more and less diverse replay buffers is slightly smaller than before. This could be due to a smaller network overfitting less to the training data, resulting in a smaller difference between better and worse generalising strategies. However, we argue that our main results with the larger network imitate a more realistic scenario in which the network architecture isn't finely tuned to the complexity of the problem. In this case, the network has a higher chance of overfitting to the training data and the positive effects of a diverse buffer could be larger. 

\end{document}